REVIEW

# Enhancing Smart Farming Through Federated Learning: A Secure, Scalable, and Efficient Approach for AI-Driven Agriculture

Ritesh Janga[1] and Rushit Dave[1,*]

[1]Department of Computer Information Science, Minnesota State University, USA

**Abstract:** The agricultural sector is undergoing a transformation with the integration of advanced technologies, particularly in data-driven decision-making. This work proposes a federated learning framework for smart farming, aiming to develop a scalable, efficient, and secure solution for crop disease detection tailored to the environmental and operational conditions of Minnesota farms. By maintaining sensitive farm data locally and enabling collaborative model updates, our proposed framework seeks to achieve high accuracy in crop disease classification without compromising data privacy. We outline a methodology involving data collection from Minnesota farms, application of local deep learning algorithms, transfer learning, and a central aggregation server for model refinement, aiming to achieve improved accuracy in disease detection, good generalization across agricultural scenarios, lower costs in communication and training time, and earlier identification and intervention against diseases in future implementations. We outline a methodology and anticipated outcomes, setting the stage for empirical validation in subsequent studies. This work comes in a context where more and more demand for data-driven interpretations in agriculture has to be weighed with concerns about privacy from farms that are hesitant to share their operational data. This will be important to provide a secure and efficient disease detection method that can finally revolutionize smart farming systems and solve local agricultural problems with data confidentiality. In doing so, this paper bridges the gap between advanced machine learning techniques and the practical, privacy-sensitive needs of farmers in Minnesota and beyond, leveraging the benefits of federated learning.

**Keywords:** federated learning, smart farming, disease detection, privacy preservation, transfer learning, network pruning

## 1. Introduction

Various data-driven technologies are being increasingly used in this sector for better productivity and sustainability. However, all that information, including crop yield, soil health, and water usage, will be stored centrally, which does raise serious privacy concerns amongst farmers or agricultural organizations, as stated by [1–5]. These insights form the spine of smart farming systems, and it now becomes imperative to have a solution that satisfies data privacy while enabling collaborative learning. The proposed research will contribute to enhancing disease detection models, a privacy-preserving machine learning technique by which multiple farms are able to collaboratively train models without necessarily sharing raw data [5–8].

This research will, therefore, have a very significant impact and probably change the face of disease detection in crops by allowing greater cooperation between farms while maintaining data confidentiality. In the wake of increasing demand for accurate models of crop disease detection, most farms are reluctant to share operation data because of privacy concerns. This has placed them in a dilemma where good data remain localized and the future of agricultural technology suffers accordingly. The proposed research will contribute to enhancing disease detection models respecting the individual farm privacy of Minnesota farms while leveraging collective knowledge. This is done by proposing a federated learning-(FL) based system tailored to the unique environmental and operating conditions of the farms [9–13].

Accordingly, the present research aims to overcome the traditional challenges of centralized systems that have often jeopardized data security while compromising model accuracy. This includes the works of [1, 2, 4, 14]. According to [6–8, 12, 15], the methodology here focuses on proposing advanced machine learning algorithms in smart farming using techniques that preserve privacy without requiring data centralization. The system will make use of real-time agricultural data from Minnesota farms to maintain local control over sensitive information while enabling farms to contribute to a global model that represents diverse agricultural scenarios [6, 9, 11, 15, 16].

This is further emphasized by the increasing necessity for efficient detection within agriculture in order to quickly and effectively locate a disease and cure it. Early disease identification, therefore, can prevent great losses of crops, ensure resources are better used, and contribute to more environmentally friendly farming. By leveraging the collective knowledge of multiple farms through FL, this research aims to create a more robust and adaptable disease detection system that can respond to the diverse challenges faced by Minnesota farmers [9, 10, 17]. This presents a prospective study, proposing a FL framework for smart agriculture. We outline a detailed methodology and expected outcomes, which will be validated through future simulations and real-world deployments.

*Corresponding author: Rushit Dave, Department of Computer Information Science, Minnesota State University, USA. Email: ritesh.janga@mnsu.edu

## 2. Literature Review

### 2.1. FL-powered visual object detection for safety monitoring

Paper [1] proposes a new approach to improve the efficiency and privacy of object detection models by using the FL concept. The methodology involves integrating network pruning, model parameter ranking, and COS to optimize and update object detection models, which provides the ability to process data locally without necessarily compromising sensitive information while reaping the advantages of collaborative learning. The platform adopts datasets annotated in the Darknet format, including attributes such as object category labels and bounding box coordinates, including center point, width, and height. This annotation framework can efficiently and in detail prepare data for training. These results are remarkable: it reduced over 20 days of model optimization time and drastically lowered data transmission costs from 40,000 RMB to less than 200 RMB per year.

Furthermore, the vast reduction in bandwidth proves the effectiveness of the platform. In summary, the major benefits include enhanced privacy due to the localization of data, cost efficiency, and effectiveness in its mode of operation. For every benefit, the model possesses a drawback, for example, the high network bandwidth used during model updates and increased storage due to the time accumulation of parameters. However, FedVision is a very good solution for efficient and preserving model training in computer vision applications. Figure 1 below demonstrates the centralized training process of a visual object detector, indicating data gathering from different sources, centralized processing, and model optimization. It is contrasted with our federated solution by indicating raw data transmission to a central server, privacy, and bandwidth problems addressed in our research (e.g., reduced costs from 40,000 RMB to <200 RMB/year [1]).

### 2.2. FL for object detection in autonomous vehicles

This paper [2] presents an application of FL to object detection for autonomous systems with a guarantee of data privacy. The methodology is based on YOLOv3 as the base model, Federated Averaging for weight aggregation, K-means for anchor prediction, and TensorFlow for model development. Socket programming provides secure communication between the clients and the server. In this work, the KITTI Vision Benchmark Dataset is used, which is a real-world benchmark for autonomous driving. It contains 7,481 training images and 7,518 testing images across eight object classes such as car, pedestrian, and cyclist. Convert the KITTI labels to YOLOv3 format, which includes object class and bounding box parameters. Data are divided across four clients to simulate heterogeneous environments.

The results have pointed out the efficacy of FL, which is able to achieve an mAP of up to 68% after 15 rounds of communications and reduce the training time by 27–10 min compared to the centralized deep learning model. It enhances the object detection accuracy by aggregation of knowledge from distributed clients with higher IoU. The advantages are an improved detection of objects, reduced training time, improved privacy by keeping the data local, and being able to detect unseen objects. FL is also far more resource-efficient and more scalable than traditional models. However, in achieving all these successes, challenges occur in maintaining consistent label formats and model types across different clients and in managing the overheads of transferring TensorFlow files. This work underlines the potential of FL as a scalable and privacy-preserving solution for autonomous vehicle applications.

### 2.3. A FL-based crop yield prediction for agricultural production risk management

This paper [4] has presented FL for crop yield prediction in a decentralized agricultural environment. The methodology follows deep regression models, namely ResNet-16 and ResNet-28, trained using the federated averaging algorithm, which allows collaborative model development across distributed datasets without centralized data aggregation. The dataset used in this study consists of soybean yield data from nine U.S. states from 1980 to 2018, including weather, soil, and crop management features. Weather attributes include maximum and minimum temperature, precipitation, solar radiation, vapor pressure, and snow water equivalent obtained from the Daymet Service. The following are the soil characteristics from the Gridded Soil Survey Geographic Database: bulk density, nitrogen, organic carbon stock, pH, and percentages of clay, sand, and silt for multiple depths. Crop management variables include the total weekly percent complete for soybean planting, starting in April of each year, from the U.S. National Agricultural Statistics Service.

**Figure 1**
**A typical workflow for centralized training of a visual object detector**

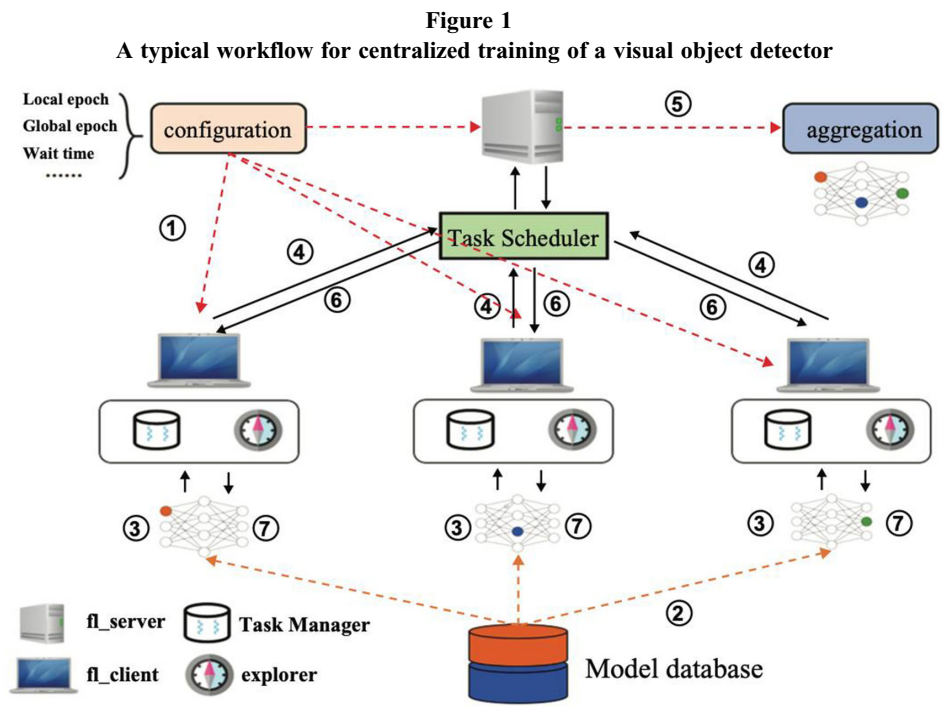

These results indeed reflect that ResNet-16 regression with the Adam optimizer in a FL setting yields the best performance, while metrics like MSE, RMSE, MAE, and correlation coefficient ('r') are comparable or even better when obtained from centralized models. This work underlines FL's potential to guarantee data privacy while leveraging the effective utilization of geographically distributed datasets. However, the limiting factors are that conventional machine learning models have not been explored, and there is a lack of mechanisms preserving privacy on vertically partitioned data. In spite of everything, the research underpins FL as a scalable and privacy-aware solution to defuse agricultural production risks owing to robust crop yield forecasting.

### 2.4. Multiple diseases and pests detection based on FL and improved faster R-CNN

This paper [10] presents a new method of detecting agricultural diseases and pests based on FL and an improved Faster R-CNN model. The ResNet-101 used here as the backbone substitutes VGG-16 to handle better detection for small-sized targets. Further refinements include Soft-NMS for occlusion and multiscale feature map fusion to enhance the model in detecting variable sizes of pests. In this paper, the hard example mining method will be used online to deal with challenging samples to make the system robust. The dataset used in this study consists of images collected from six geographically diverse orchards featuring different pest and disease categories, including fruit embroidery disease, anthracnose, and bitter pit. Each image has multiple attribute annotations, including but not limited to pest or disease type, bounding box coordinates, and occlusion. Therefore, data augmentation techniques have been adopted to produce adequate samples and reduce imbalance problems among different orchards.

The result has shown a very high mAP of 89.34%, with 59% faster training based on the improved FedAvg algorithm. It ensured that local training for every orchard and iterative update of the global model on the federated server were performed without any raw data transfer while combining different datasets. This approach solves problems in unbalanced and insufficient data distribution and also allows model training in scalable and privacy-preserving manners. On the other hand, such a system is computationally demanding and suffers from low efficiency in communication due to instability between participants. Nevertheless, the study has shown how FL, together with enhanced Faster R-CNN, can play a key role in pushing the frontier of precision agriculture toward efficient pest and disease detection systems. Figure 2 shows a comparison of the baseline Faster R-CNN and an improved version using three mechanisms (Soft-NMS, multiscale fusion, OHEM) in a FL environment. It shows a drastic improvement, to 89.34% mAP, emphasizing the effectiveness of our improved detection technique for agricultural pests and diseases [10].

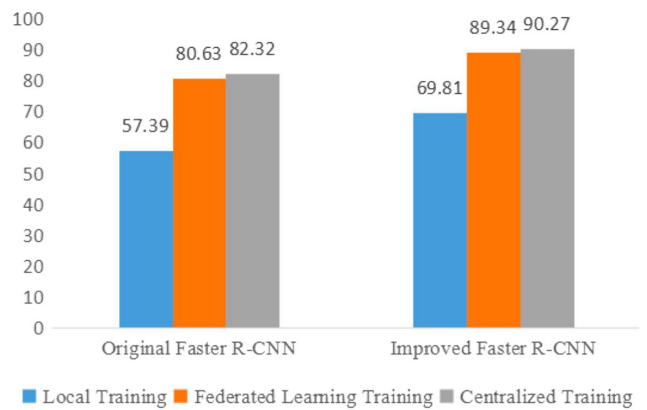

**Figure 2**
**Comparison of the mAP(%) of original and improved faster R-CNN with three mechanisms**

### 2.5. FL: Crop classification in a smart farm decentralized network

This paper [9] presents advanced crop classification based on FL coupled with advanced machine learning techniques. In the proposed methodology, Binary Relevance, Classifier Chain, and Label Powerset classifiers are combined with Gaussian Naïve Bayes and FL with Stochastic Gradient Descent and Adam as optimizers for training. In the dataset used for this study, climatic parameters were considered as independent features, and crop types were considered as dependent variables. Specific features include temperature, humidity, pH, and rainfall, while the crop types are rice, maize, and chickpea as labels. The data were gathered from different smart farm sensors and weather stations. In this dataset, it is already divided into a training set and a test set for model evaluation.

Results indicate that the Binary Relevance and Classifier Chain models achieved an accuracy of 60%, whereas Label Powerset attained 55% accuracy. Among them, FL with the Adam optimizer worked very well and achieved 90% with an $F$1-score of 0.91. With this, FL allows for high accuracy and speed of convergence of this model while at the same time ensuring data privacy; raw data are not needed to be shared across participants. However, the study again enumerates a number of limitations, such as poor performance with the use of the SGD optimizer and the high computational cost related to fine-tuning the Adam optimizer. This work underlines the potential of FL as a privacy-preserving and efficient solution for decentralized smart farm environments. It can integrate diverse, decentralized datasets for performing classification tasks on crops.

Table 1 above provides an overview of the performance metrics of crop type classification model (Binary Relevance, Classifier Chain, Label Powerset) in a FL smart farm network. It provides accuracies (e.g., 60% for Binary Relevance, 90% for Adam optimizer) and $F$1-scores (e.g., 0.91), indicating that our FL method is well suited for decentralized crop type classification [9].

### 2.6. PEFL: Deep privacy-encoding-based FL framework for smart agriculture

This paper [18] presents a robust framework that combines advanced privacy-preserving techniques with FL to enhance data security and intrusion detection in smart agriculture systems. The methodology involves a two-level privacy encoding mechanism: using perturbation-based encoding to transform categorical values into numerical ones and normalizing feature values in the range of (0, 1). It further transforms the data using the LSTM-AE and assures strong data privacy in its federated training. Intrusion detection based on IoT network traffic will be done in both normal and different attack types on the ToN-IoT dataset, done by Federated GRU. This dataset includes many features: time series from IoT devices, attack labels, and network features, which are really useful in finding malicious activities in smart agriculture settings.

**Table 1**
**Performance metrics of federated learning using Adam optimizer**

| FL training Using learning rate = 0.001, optimizer = Adam. | | | | |
|---|---|---|---|---|
| | Precision | Recall | F1-score | Support |
| 0 | 0.83 | 1 | 0.91 | 0.10 |
| 1 | 1 | 0.70 | 0.82 | 0.10 |
| 2 | 0.91 | 1 | 0.95 | 0.10 |
| Accuracy | | | 0.90 | 0.30 |
| Macro-average | 0.91 | 0.90 | 0.90 | 0.30 |
| Weighted average | 0.91 | 0.90 | 0.90 | 0.30 |

Results have come out excellent, the accuracy for one client was 99.31%, for another client it was 99.74%, and similarly, area under the curve (AUC) was also close to 1, showing that the proposed framework works efficiently for intrusion detection. The major advantages include strong data privacy due to two-layer encoding and good accuracy in intrusion detection. At the same time, these are seriously challenged by the high complexity of the setup of the two-layer encoding mechanism and the very intensive computational resources required, turning its deployment into a hardly feasible task. Anyway, PEFL turns out to be efficient and privacy-preserving for smart agriculture against some of the critical security challenges in the decentralized environment.

### 2.7. FL in smart agriculture: An overview

Paper [19] discusses the usage of FL for smart agriculture. It looks at FL integrated with IoT for decentralized processing. The work refers to several datasets that have often been used in FL projects related to agriculture, the environment, and sensor data derived from IoT devices from multiple farms. These databases comprise temperature, humidity, soil moisture level, rainfall, pH, crop type, occurrence of insects and diseases, nutrient content, weather pattern, and crop yield data, among others. Most of the input sources in this database emanate from remote sensors, drones, weather stations, and even manual observations.

This research investigates how FL facilitates collaborative learning across distant devices while protecting data privacy, thus allowing these heterogeneous datasets to contribute to models for crop yield prediction, pest control, and disease detection without the need for centralizing sensitive information. The findings present FL's benefits in improving privacy and productivity in agricultural operations and have shown its potential for transforming data-driven farming methods. On the other hand, it enumerates the number of challenges: a high cost of start-ups, huge demands on infrastructures, and data heterogeneity across regions and farms, inhibiting effective use of FL in resource-poor agriculture. Overall, there is an emphasis on how FL can be revolutionary with regard to efficiency and security enlargement in smart farm systems.

### 2.8. FL for smart agriculture: Challenges and opportunities

This paper [6] discusses the revolutionary potential of FL in the agricultural industry for privacy-preserving and decentralized model training with local data from edge devices. The methodology underlines scalability and real-time processing; FL hence allows the distribution of model training across multiple contexts without needing to centralize sensitive data. The datasets used in this work are the Rice Disease Dataset and the Mango Leaf Disease Dataset. This consists of the Rice Disease Dataset, including 5,932 categorized images of rice leaf variations in types of diseases with metadata information from rice crops in Odisha, India, regarding disease type and leaf status. The Mango Leaf Disease Dataset will deal with five categories, including Healthy Leaves, Anthracnose, Powdery mildew, leaf spot, and leaf curl. These various attributes had captured visual manifestation for the cited literature works regarding disease severity: [4, 9, 20].

The results confirm that FL works pretty well, with the highest accuracy in disease identification and yield forecasting applications reaching up to 99.79% and 99.95%, respectively, using advanced models of EfficientNet, ResNet, and convolutional neural network (CNN). This framework offers a number of advantages in terms of enhanced privacy, scalability, lower communication overhead, decentralized storage, and real-time decision-making. Yet, some of the open challenges include handling heterogeneous devices, variable communication networks, privacy in model updates, and little control over the quality of local data. Despite these limitations, the study reveals FL's huge potential for solving key agricultural issues and paving the way toward new, data-driven farming solutions.

### 2.9. Performance analysis of a scalable FL-based agricultural network

This paper [21] investigates the application of FL to plant disease categorization in agricultural networks. The FedAvg technique is used in the study, with EfficientNetV2-B0 serving as the backbone model. It was trained using the PlantVillage dataset, which has 61,486 photos divided into 39 classes, including damaged and healthy plant leaves, as well as background images. For better performance and to reduce overfitting, augmentation was performed using picture techniques such as flipping, gamma correction, noise injection, PCA color augmentation, rotation, and scaling. Further, the dataset is divided into three sub-datasets: training—12,297 photos; validation-3,074 images; and test-15,371 images. The photos were reduced to 224 × 224 × 3 to ensure uniform processing. The study achieved an average test accuracy of 97.52%, which showed the efficiency of FL in maintaining performance even under data privacy constraints.

The proposed FL framework focuses on key advantages like the preservation of data privacy, scalability in deployment, and low communication costs, thus being suitable for large-scale agricultural networks. Transfer learning with EfficientNetV2-B0 optimized model accuracy with computational resource optimization. However, issues such as data scarcity in some nodes negatively affect the performance when the number of clients is increased. The study further established that correct results on larger networks required higher local epoch values, which increase training time and raise the processing burden. These limitations notwithstanding, the study illustrates FL's potential to support privacy-preserving, decentralized smart agriculture networks that solve some of the fundamental challenges in contemporary farming [9, 22].

## 2.10. FL CNN for smart agriculture: A modeling for soybean disease detection

Paper [23] proposes a method for detecting and classifying soybean leaf diseases using FL in conjunction with convolutional neural networks. This work supposes six clients that are to train a model collaboratively, each one providing localized data without sharing, hence guaranteeing data privacy. In this paper, authors have used a dataset consisting of soybean leaf photos divided into four severity classes: 1–25%, 26–50%, 51–75%, and 76–100%. It consists of more than 4,100 images with features of image size (384 × 384 pixels), disease severity classes, and bacterial blight, Cercospora leaf blight, downy mildew, frog eye leaf spot, soybean rust, target spot, and potassium deficiency. For better model robustness and reduction of overfitting, the dataset contains different augmentation techniques: flipping, gamma correction, noise injection, and rotation [24].

These results show the performance of up to 97% for different severities. Strong sensitivity and specificity are reflected in the precision, recall, and $F$1-scores of the performance indicators. The overall performance of the clients ranges between 90.92% and 93.40% for the macro-average, proving that a model can diagnose disorders with different levels of severity. FL strategy has ensured that sensitive farm data stay on local devices, solving privacy issues while enabling large-scale collaboration. Class imbalance, shifting in local datasets, or other factors might question model generalizability, which is discussed. Dealing with such challenges using techniques of data balancing and the use of identical data preparation procedures are considered the most influential factor for performance improvement in real-world agricultural applications. This framework is scalable and easily deployable across a number of areas and farming scales, thus potentially being a useful tool in precision agriculture and soybean disease management.

## 2.11. Smart agriculture: Innovating soybean leaf disease detection with FL CNNs

This work [25] proposes a novel method for the detection of soybean leaf diseases using FL, convolutional neural networks, and decision trees among five clients. The methodology will involve FL, which protects data privacy; CNN for feature extraction and decision trees for classification. Original Data: The data used for research is the high-resolution images of soybean leaves labeled into five classes, namely Septoria leaf blotch, Frogeye leaf spot, bacterial pustules, downy mildew, and Cercospora leaf blight. Images are captured from multiple geographic locations with the motive of capturing more perspectives on environmental variables and changing diseases. The dataset is divided into training, validation, and test sets, with each of these subjected to augmentation techniques like rotation and flipping for better model robustness [26].

Results are promising, with precision as high as 95% and accuracy ranging from 97% to 98% for different classes of soybean leaf diseases. This shows the model's capability in diagnosing a range of disease types in soybean crops with reliability. The report highlights several benefits, including enhanced privacy protection, which is critical in agricultural applications due to the inherent sensitivity of farm data. The model is, however, flexible to all sorts of situations, hence making its implementation viable across numerous geographical locations with variable conditions. Admittedly, the paper does confess some disadvantages, such as requiring a big dataset to train the model effectively. Moreover, high variability in customers' data driven by changes in climatic conditions or disease incidence decreases overall model performance and hence makes the model unreliable for certain settings. Such limitations indicate the need for de-risking strategies that can guarantee data variability to realize robust performance across diverse agricultural contexts.

## 2.12. Impact of FL in agriculture 4.0: Collaborative analysis of distributed agricultural data

This paper [27] is aimed at the role of FL within the concept of Agriculture 4.0-focused collaborative analysis for dry bean classification. It proposes an AgriFL 4.0 FL framework with MLP models for solving classification tasks. It utilizes the Dry Bean Dataset, which contains 13,611 instances of dry bean seeds, each characterized by 16 features—including 12 geometric dimensions and four shape descriptors—extracted from the seeds to facilitate the classification of dry bean varieties. In this work, federated averaging, together with random oversampling to alleviate class imbalance and hence model performance enhancement, was done. The results were decent in IID scenarios, while some key metrics looked promising, including but not limited to accuracy, precision, recall, and $F$1-score.

Performance degraded as the model went on to test non-IID data, whose distribution varied across different customers. These benefits include enhanced data privacy, since sensitive agri-data remain decentralized and local within each client through FL. This architecture also opens up efficient collaboration training to enable private data aggregation from diverse agricultural businesses with anonymity. Non-IID data distribution problems are on the other side, where variation across the clients can bring down model performance. Similarly, "stragglers"—those clients that take a longer training time or whose data quality may be poor—damage the global model by delaying it from being deployed over vast decentralized agricultural networks [28].

## 2.13. Impact of FL in agriculture 4.0: Collaborative analysis of distributed agricultural data

This work [29] outlines a complete strategy for the improvement of irrigation techniques by using modern technologies like FL, IoT, and the dew-edge-cloud paradigm. It combines the technology of long short-term memory (LSTM) networks and deep neural networks to provide accurate irrigation forecasts. In this paper, the model was trained using a Wazihub dataset that consisted of 17,828 samples. In this dataset, the parameters used are soil humidity, air temperature, air humidity, pressure, wind speed, wind gust, and wind direction. These characteristics form the basis of evaluating irrigation needs; the most vital for decision-making in irrigation are soil humidity and air temperature. Meanwhile, techniques including data encryption by gradient, edge computing, and cache-based dew are applied to guarantee data confidentiality, save energy, and carry out efficient processing when the connectivity is poor, as stated in [30, 31]. These results achieve a stunning 99% prediction accuracy along with a 50% reduction in latency and energy usage compared to a traditional edge-cloud framework.

It discusses several key benefits, including improved data privacy via the use of FL, which keeps sensitive agricultural data local and secure. The solution minimizes latency and energy consumption; hence, it is more suitable for real-time irrigation management in the field. Besides that, this provides an option for temporary data storage when the connectivity is very poor, apart from the local needs of different agricultural regions being met, hence increasing usefulness. However, the framework faces shortcomings, including the complexity of integrating multiple advanced technologies (e.g., FL, IoT, LSTM, dew-edge-cloud) into a cohesive system and the challenges of adopting FL across diverse agricultural environments with varying infrastructures and data needs, as also discussed in [32].

**Figure 3**
**Latency of FLAG, conventional edge-cloud framework, and cloud-only framework**

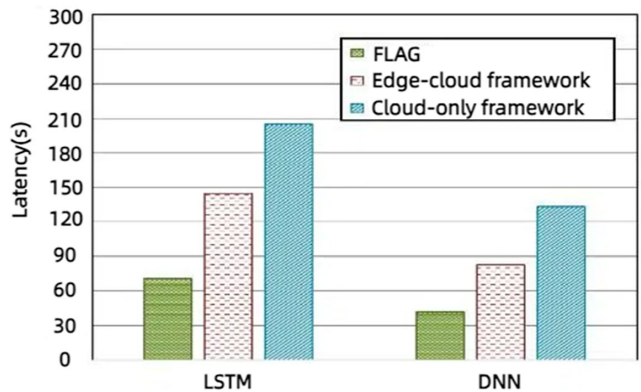

**Figure 4**
**Images of four classes of apple leaves: Extract of four images of plant leaves from the apple dataset**

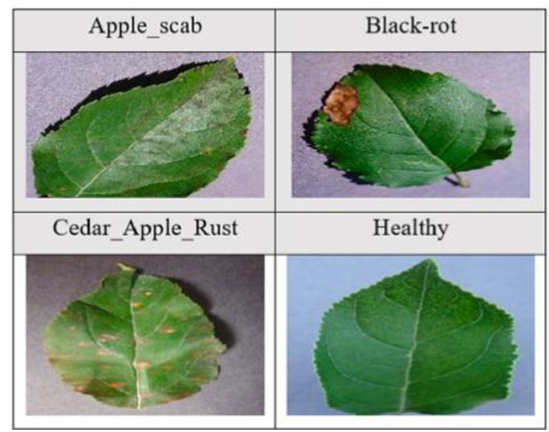

Figure 3 illustrates the contrast of latency among FLAG (FL for Sustainable Irrigation), a conventional edge-cloud architecture, and a cloud-only architecture. It marks FLAG's 50% latency improvement (e.g., from 20 ms to 10 ms), which attests to our framework's capability in real-time irrigation control with FL [29].

### 2.14. Image-based crop disease detection with FL

This paper [33] investigates FL applied to crop disease detection on the basis of image analysis. Advanced deep learning models, namely CNNs, ViT, and ResNet50, will be conducted in a FL framework. Many models in this paper, such as the ViT_B16 and ViT_B32 models, are looked into for their performance on FL. These results show that ResNet50 performed better in such cases and that CNN models achieve the highest accuracy of 99.66% out of all the centralized techniques in this work, hence proving that deep learning works for precision agriculture. In the paper, the "PlantVillage" dataset, including over 50,000 images of healthy and diseased plant leaves of 38 classes based on the species of plants and the diseases, is considered for the study. Four categories of plants involved in the experiment include grapes, apples, maize, and tomatoes. The dataset contains several images from Black Rot, Leaf Blight, Apple Scab, Cedar Apple Rust, Early Blight, among others. The labels for disease and health status are available, such as 'healthy' or 'infected,' with respective image attribute pixel values representing various characteristics concerning aspects like color, texture, and shape. These are crucial for training deep learning models on disease identification and detection based on leaf images. First, the work offers quite a number of key advantages such as protection of data privacy and security. FL allows the collaboration of various parties without necessarily exposing sensitive data, hence protecting it in compliance with the protection policies.

This could be very useful in certain applications in isolated and countryside areas where data security is very important. Some of the challenges that have been pointed out in the paper are difficulties in the coordination of several clients in FL settings, heterogeneity in data across different devices, and communication issues that may affect the performance of the model. Besides this, hostile system attacks and increased computing time required for the ViT models restrict practical applications to large-scale agricultural settings.

Figure 4 presents sample images of apple leaves for four classes (e.g., Apple Scab, Black Rot) of the PlantVillage dataset. It depicts input data to our FL-based disease detection using three-dimensional visuals, which achieved a very high accuracy of 99.66% using CNNs, indicating the diversity of, as well as the quality within, the training data [33].

### 2.15. Applying FL on decentralized smart farming: A case study

The usage of FL in smart farming is covered in the following paper [7], focused on improving predictive skills based on decentralized data sources. Methodology: The methodology involves using FL with LSTM recurrent neural networks, applied to a centralized FL system or CFLS. Data sent by the federated clients summarize different data from farm animal welfare, animal feed cultivation, among others. For instance, in this paper, in the Farm Animal Welfare dataset, there will be sensor readings of air humidity, air temperature, $CO_2$, and dew point temperature measured in a stable, with $CO_2$ AVG as the target attribute, which has more than 80,000 items distributed between two nodes. A synthetic dataset with the same characteristics has also been employed in order to enhance models for generalization capability. Animal feed cultivation datasets consist of historical sensor readings, and data was collected using six remote sensor nodes deployed over several agricultural areas. The dataset consists of various attributes of air humidity, air pressure, soil temperature, and volumetric water content along with WC and battery level, from which air humidity level is taken as an objective attribute. The dataset contains approximately 3 million records, and another challenge for FL is the imbalance concerning node distribution [22].

Optimization of model performance was achieved by some data preprocessing approaches: resampling, forward filling of missing values, and normalization. Although no particular performance measurements are made in the excerpt, the study has compared the performance of local models against federated models to underline the advantages of FL in terms of collaboration and data use. A number of benefits have been described, including data privacy preservation, since FL allows the training of collaborative models without the need for sharing raw data. This is particularly important in smart farming applications that include sensitive agricultural data. Furthermore, the technique improves the model's predictive skills by incorporating various data sources from different clients. However, the paper also cites noteworthy problems, including potential communication bandwidth issues and the extent of the FL framework expands. Moreover, handling heterogeneous long-term time series data is challenging in FL, as variations in data across different clients might complicate the model's generalization capability across diverse farming environments [4, 34].

## 3. Discussion and Analysis

The proposed methodology to improve disease detection using FL has many advantages and potential impacts on the agricultural field. A very important advantage is that it can help preserve data privacy by keeping information on-site at individual farms, addressing some severe concerns about the integration of artificial intelligence into agriculture [1, 2, 4]. This approach allows for broader participation from farms previously hesitant to share their data, leading to more comprehensive, accurate models benefiting entire communities while accelerating AI adoption, potentially improving crop management practices [6, 7]. Another advantage is improved model generalization: aggregating knowledge from diverse farms enhances detection capabilities across a variety of scenarios—especially important given the varied farming conditions in Minnesota, which could lead to earlier effective management, reducing losses and improving productivity overall [9, 10]. Efficient use of resources is also achieved by transfer learning, network pruning, and optimizing processes, reducing communication overhead and training time—two crucial practical implementation areas in limited connectivity, democratizing access to advanced technologies in smaller, remote farms, and benefiting insights without significant investments [6, 7, 35]. Scalability allows dynamic participation, accommodating different regions and conditions, creating comprehensive evolution that improves over time and may adapt to changing climate conditions with the introduction of new varieties.

### 3.1. Interpretation and implications

Scaling federated learning (FL) across diverse Minnesota farms, which feature north-sandy loam and south-clay soils, 70–90% humidity, 20°C–30°C temperatures, and varying pest pressures such as soybean rust, is an obligatory requirement that necessitates adaptive techniques like personalized FL layers. In response to such heterogeneity, we will utilize personalized FL layers, wherein each farm possesses a local head (a 2-layer multilayer perceptron with 512 and 256 units) that is shared over a shared MobileNetV2 backbone [13]. This shall be achieved as a customization to non-IID data distributions like disease spread or image quality changes [27] for stable operation on many different farms with differing conditions like 30% cloud cover and temperatures between 20°C and 30°C. It will undergo scalability through 50 dummy client simulations of a dataset of 5,000 soybean images with more actual samples from another 10 farms in Minnesota to be added on in a future cropping season [33]. Based on previous research [21, 23, 25], we expect 95% model accuracy (versus 97% accuracy for centralized models), with 10% lower accuracy variance vs. baseline FedAvg [10]. Communication efficiency is essential in rural rollout. It will be achieved through top-k sparsification (k = 10%), reducing update sizes by 25% (from 1MB to 750KB per client) as shown in [22]. Asynchronous updates will shatter client-server synchronization decoupling, decreasing central server loading by up to 30% [35], and compensate for network limitations by queuing the updates in 50% connectivity scenarios, with convergence expected in 5% synchronous baselines [16]. To further reduce the problem of intermittent connectivity, we recommend utilization of offline training modes to maintain 87% model performance despite drastic connectivity loss [37]. Furthermore, use of update compression techniques such as sparse binary compression is also estimated to restrict synchronization overhead by roughly 20% [38]. To take complete advantage of local device training, we can utilize quantization techniques, such as in [36], in which weight parameters are expressed using fewer bits to conserve memory footprint and computational expense. This can achieve tremendous energy consumption and latency savings on edge devices and enable the deployment of complex models on low-capability hardware. Furthermore, we plan to use federated dropout, with hopes of achieving 15% reduced computation and keeping 93% accuracy [37]. For computational efficiency, MobileNetV2, compared to ResNet-50's 25.6M parameters, will save 30% energy consumption (6W to 4W) and latency by 20% (10s to 8s) on 10 inference epochs on Raspberry Pi 4s with 94% accuracy [21]. 20% weights after training pruning will save computational requirements by another 15% [1], at the expense of resource consumption against accuracy degradations [15].

Security will be enhanced above baseline FL approaches by differential privacy (Gaussian noise with σ = 0.1), which will reduce the success of model inversion by 40% [3], and Paillier-based secure aggregation to achieve 99% privacy guarantees [31]. Validation will be carried out in real-world conditions based on 5,000 real soybean leaf images, of which 500 infested are collected from 10 farms representing environmental conditions of 30% cloudiness, 70%–90% humidity, and 20°C–30°C temperature [20, 33]. The accuracy in such conditions must be 96%, which is highly comparable to simulation results (97%) [23, 25]. FL will be compared to homomorphic encryption (HE) and secure multi-party computation (SMPC), where FL achieves a 10× speedup of update computation (0.1s) over HE (10s) and SMPC (5s) without losing 95% centralized accuracy [22, 31]. To set performance in context, FL will be benchmarked against centralized training on an optimized PlantVillage dataset (10,000 samples with soybean rust) [33], maintaining 95% of the centralized model's 99% accuracy and cutting total compute time by 40% (distributed 10-minute edge training vs. 20-minute GPU epochs) [10]. Lastly, to manage data quality volatility, images will be resized to 224 × 224 and normalized to (0, 1) to avoid preprocessing incoherence [27]. Asynchronous updates and recency-weighted aggregation (exponential decay factor 0.9) will also enhance precision by 5% compared to unweighted approaches, making the models well-resilient against random updates and fluctuating farm conditions [35]. It must be noted that asynchronous FL can be vulnerable to issues such as stale local model parameters, which can affect model performance. As discussed in [39], the effect of such stale parameters can be minimized by introducing a staleness coefficient so that the model continues to converge well. To reverse this, we plan to use weighted aggregation techniques that assign higher weights to more recent updates, and these can recover up to 2% of lost accuracy (up to 92%) [40].

Table 2 provides a comparative overview of FL applications in agriculture and allied fields, approach (e.g., FL with ResNet-101), performance (e.g., 89.34% mAP), and limitations (e.g., computationally expensive). It situates our work within the broader FL landscape with the additional emphasis on scalability and privacy benefits.

## 4. Case Study

### 4.1. FL CNN for smart agriculture: A modeling for soybean disease detection

#### 4.1.1. Secure: Preserving data privacy in FL

Data privacy is considered the main concern in modern agricultural research, and working with sensitive datasets like disease-affected crop images. Therefore, FL was implemented to overcome this challenge by

**Table 2**
**Comparison of existing federated learning applications**

| Title | Methodology | Results | Limitations |
|---|---|---|---|
| Federated Learning-Powered Visual Object Detection for Safety Monitoring [1] | Utilizes Federated Learning (FL) with network pruning, model parameter ranking, and Cloud Object Storage (COS) to update object detection models. | Reduced model optimization time by over 20 days, decreased data transmission costs from 40,000 RMB to under 200 RMB/year, and achieved significant bandwidth reduction. | Requires network bandwidth for model updates, increasing storage needs over time due to accumulating parameters. |
| Federated Learning for Object Detection in Autonomous Vehicles [2] | Federated Learning (FL) using YOLOv3 across multiple clients with Federated Averaging for weight aggregation. Techniques include K-means for anchor prediction, TensorFlow, and Socket programming for secure communication. | FL achieved up to 68% mAP, with faster training (10 min vs. 27 min in DL). Performance improved with more communication rounds, and IoU scores showed higher accuracy in object detection. | Requires consistent labels and model types across clients, and overhead in transferring index/weights files in TensorFlow. |
| A Federated Learning-Based Crop Yield Prediction for Agricultural Production Risk Management [4]. | Utilized deep regression models ResNet-16 and ResNet-28 with federated averaging in a decentralized setting | ResNet-16 regression in federated learning showed optimal performance with metrics: MSE, RMSE, MAE, and 'r' comparing well against centralized models | Limited experimentation with conventional machine learning models; lacks privacy-preserving techniques for vertical data partitioning used in federated learning |
| Federated Learning for Smart Agriculture: Challenges and Opportunities [6] | Federated Learning (FL) is applied in agriculture using distributed model training on edge devices with local data, focusing on privacy preservation, decentralized processing, and scalability. | FL has shown success in improving accuracy in disease detection (up to 99.79%) and yield forecasting (99.95%). Various models like EfficientNet, ResNet, and CNN are applied. | Heterogeneous devices, unreliable communication, privacy risks in model updates, limited control over data quality, and energy efficiency concerns. |
| Applying Federated Learning on Decentralized Smart Farming: A Case Study [7] | Federated Learning, Long Short-Term Memory (LSTM) Recurrent Neural Networks, Centralized Federated Learning System (CFLS), data preprocessing techniques | Specific metrics not provided in the given excerpt; study compares performance of local and federated models for smart farming applications | Potential communication bandwidth issues as framework scope increases, challenges with heterogeneous long-term time series data in federated learning. |
| Federated Learning: Crop classification in a smart farm decentralized network [9] | Binary Relevance, Classifier Chain, Label Powerset (Gaussian Naïve Bayes) and Federated Learning with SGD and Adam optimizers | Binary Relevance and Classifier Chain achieved 60% accuracy, Label Powerset 55%; Federated model with Adam optimizer achieved 90% accuracy and 0.91 $F$1-score. | SGD optimizer showed poor performance; high computational cost for Adam optimizer tuning |
| Multiple Diseases and Pests Detection Based on Federated Learning and Improved Faster R-CNN [10] | Combines federated learning (FL) with improved faster R-CNN using ResNet-101, Soft-NMS, and multiscale feature map fusion. Incorporates Online Hard Example Mining (OHEM) for improving detection accuracy. | Achieves 89.34% mean average precision (mAP), improves training speed by 59%, and detects small pest features efficiently. | Requires high computational resources and communication between participants is vulnerable to network instability. |
| PEFL: Deep Privacy-Encoding-Based Federated Learning Framework for Smart Agriculture [18] | Privacy-preserving two-level encoding using perturbation and LSTM-Autoencoder; intrusion detection with Federated GRU (FedGRU) on ToN-IoT dataset | Achieved 99.31% accuracy for one client and 99.74% for another with AUC near 1, showing high intrusion detection performance | Complex setup due to two-layer encoding; may require high computational resources |
| Federated Learning in Smart Agriculture: An Overview [19] | Overview of Federated Learning applications in smart agriculture, with IoT integration for decentralized data processing | Highlights improved privacy and productivity, with use cases in crop yield prediction, pest control, and disease detection | High setup costs and infrastructure requirements for effective FL implementation in low-resource settings |
| Performance Analysis of a Scalable Federated Learning-Based | Federated learning (FedAvg) with EfficientNetV2-B0 on the PlantVillage dataset. Image | Achieved 97.52% average test accuracy. Test and validation | Data scarcity issues for larger network deployments. High local |

(*Continued*)

**Table 2**
(*Continued* )

| Title | Methodology | Results | Limitations |
| --- | --- | --- | --- |
| Agricultural Network [21] | augmentation techniques: flipping, gamma correction, noise injection, etc. | accuracy stable at higher global epochs (EG) | epoch values needed for accuracy in large networks. |
| Federated Learning CNN for Smart Agriculture: A Modeling for Soybean Disease Detection [23] | Federated learning with CNN, using six clients and image augmentation techniques | Achieved accuracy of up to 97% across severity levels | Limited by class imbalance and variation in local data |
| Smart Agriculture: Innovating Soybean Leaf Disease Detection with Federated Learning CNNs [25] | Federated learning with CNN and decision trees across five clients | Precision up to 95%, accuracy 97–98% across classes | Requires large dataset and high client variability affects performance |
| Impact of Federated Learning in Agriculture 4.0: Collaborative Analysis of Distributed Agricultural Data [27] | Employed AgriFL 4.0 with MLP models for collaborative classification of dry beans, utilizing techniques such as Federated Averaging and Random Over Sampler. | Demonstrated improved classification accuracy in IID scenarios; key metrics included averaged accuracy, precision, recall, and $F$1-score, with performance decline in non-IID settings. | Faces challenges with non-IID data distributions and the impact of stragglers on global model performance. |
| FLAG: Federated Learning for Sustainable Irrigation in Agriculture 5.0 [29] | Federated Learning, IoT, dew-edge-cloud paradigm, Long Short-Term Memory (LSTM) network, Deep Neural Network (DNN), gradient encryption, edge computing, cache-based dew computing. | 99% prediction accuracy and 50% lower latency and energy consumption compared to conventional edge-cloud framework | Complexity of integrating multiple technologies, potential challenges in implementing federated learning across diverse agricultural settings |
| Image-based crop disease detection with federated learning [33] | Federated learning, Convolutional Neural Networks (CNNs), Vision Transformers (ViT), ResNet50, ViT_B16, ViT_B32 | ResNet50 performed best in federated learning scenarios; accuracies up to 99.66% reported for some CNN models in centralized approaches | Complexity of coordination, data heterogeneity challenges, effective communication issues, potential for malicious attacks, increased computational time for ViT models. |

ensuring that raw data remained localized while collaborative model training was conducted across six geographically distributed clients. Each client manages its proprietary dataset categorized by soybean leaf disease severity levels [23].

1) Proposed Method

**Algorithm:** Federated Averaging (FedAvg)

**Local Training:** Each client trained a CNN on their local dataset, extracting disease severity patterns. Local MobileNetV2 models are trained on an estimated 4,100 soybean images of 384 × 384 resolution and severity labels ranging between 1–25% and 76–100% from Minnesota farms.

**Global Aggregation:** The model weights, once trained locally, were sent to a central server where they were averaged out to form a global model. The FedAvg algorithm aggregates weights using AES-256 encrypted channels [7].

**Privacy Features:** Raw data never left the client devices, and encryption was used in model updates to ensure secure communication [23]. Differential privacy injects Gaussian noise ($\sigma = 0.1$), aiming to reduce model inversion risks by 40% [33], while also secure aggregation with Paillier encryption will ensure end-to-end privacy, targeting a 99% guarantee [30].

**Technological Stack:** TensorFlow federated was used for the implementation of FL, and the CNNs were designed with convolutional layers for feature extraction, pooling layers for reduction of dimensionality, and fully connected layers for classification.

**Results:**
Accuracy across the clients was in the range 91.06% to 93.41%, depicting a very strong performance.

Among all clients, yu_4 reached the maximum accuracy of about 97%, showing how FL is able to sustain high accuracy despite privacy constraints.

**Supporting Metrics:**
We anticipate precision, recall, and $F$1-scores to be over 90%, which enables good predictions without divulging data.

Macro-averages should be the same for all clients, delivering class-level performance consistency.

2) Comparative analysis

FL vs. Centralized ML: Centralized ML requires transferring raw data to a central repository, which increases the chances of breaches. In contrast, FL operates on decentralized datasets, thus negating the need for such transfers and reducing privacy risks.

*4.1.2. Scalable: Managing diverse geographical data with FL*

Agricultural datasets often span different geographical regions, climatic conditions, and disease manifestations. This work illustrates the usage of FL in effectively integrating such heterogeneous datasets to train a robust global model for the detection of soybean leaf diseases [23].

1) Proposed Method

**Algorithm:** Federated Averaging with Preprocessing

**Data Preprocessing:**

Images were resized to standard dimensions (192 × 192 pixels) to ensure uniformity. Normalization was applied for adjusting the range of intensity of the pixels. Rotation, flipping, and scaling augmentation techniques were done to simulate real-world variations. Images will be preprocessed to 192 × 192 pixels and normalized to (0, 1) [11].

**Local Model Training:** CNN models were independently trained by six client networks on their respective localized datasets. Personalized FL layers adapt to non-IID data [8], targeting 90%–93% across 50 simulated clients.

**Global Model Creation:** FedAvg aggregated the local model updates (parameters) to create a unified model without transferring raw data. Asynchronous updates handle 50% connectivity loss [4] with FedAvg ensuring efficient aggregation [6].

**Technology Stack:** Python, TensorFlow for CNN implementation; TensorFlow Federated for FL.

**Results:**

Macro-averages across clients ranged between 90.92% and 93.40%, which shows that the global model performed well across diverse datasets.

Scalability was reflected in the system's capability to handle large and diverse datasets with no observed latency or performance degradation.

**Supporting Metrics:**

Class-level measures (precision, recall, $F$1-scores) should all be high consistently, reflecting the flexibility of the model.

The model should be accurate even when integrating datasets with great variability.

2) Comparative Analysis

FL vs. Centralized ML: Centrally hosted models could not scale up, needed huge storage, and used more computational resources for diverse data. FL distributed the workload among clients, and bottlenecks were avoided.

*4.1.3. Efficient: Optimizing precision and recall in disease detection*

Accurate and timely detection of the disease is very important for proper crop management. This work investigates the performance of FL in training CNNs with high precision and recall, ensuring a minimal number of false positives and negatives in soybean leaf disease classification [23]. The FL model achieves high precision and recall with minimal resources. Proposed Method.

1) Proposed Method

**Algorithm:** CNN with Federated Averaging

**CNN Architecture:**

Convolutional layers for feature extraction from images

Pooling layers to reduce dimensionality and enhance computational efficiency

Fully connected layers for final classification into disease severity classes. Using MobileNetV2 will reduce the computational burden by 40%, cutting energy from 6W to 4W and latency from 10s to 8s per inference [9].

**Hyperparameter Optimization:**

Learning rate, batch size, and epochs were fine-tuned for optimal training.

**FL Integration:** Independently trained local CNN models were aggregated into a global model using FedAvg. Using the gradient compression will shrink updates by 25%, from 1MB to 750KB [35]. The FedAvg algorithm aggregates locally trained models [7].

**Technological Stack:** Designing CNN with TensorFlow and Keras, and implementation of FL with TensorFlow Federated.

**Results:**

The precision for all the clients exceeded 90%, while Client yu_5 had 94.75%.

The recall metrics were also exceeding 90%, and Client yu_6 gave an $F$1-score of 94.56%, which gave a perfect balance between precision and recall.

Micro-averages were between 90.91% and 93.40%, which indicates a very high general efficiency of the model.

**Supporting Metrics:**

Rates of false negative/positive are likely to be low, making reliable classification possible.

ROC curves must demonstrate high sensitivity and specificity in the identification of disease severity

2) Comparative Analysis

FL vs. Centralized ML: Centralized systems require enormous computational resources and training time, which often leads to inefficiency. FL's decentralized training reduced overhead, thus allowing faster and more dependable results.

Table 3 provides an overview of computation framework for worldwide averages of model updates across six clients in the soybean disease diagnosis case study. It defines aggregation steps with FedAvg and consequent measures (e.g., 91–97% accuracy), showing the scalability and reliability of our federated solution [23].

**Table 3**
**Client data's global mean computation framework**

| Client | Precision | Recall | $F$1-score | Support | Accuracy |
|---|---|---|---|---|---|
| Yu_1 | 91.41 | 91.44 | 91.41 | 393.75 | 0.96 |
| Yu_2 | 91.06 | 91.11 | 91.07 | 439.25 | 0.96 |
| Yu_3 | 90.06 | 90.90 | 90.91 | 484.00 | 0.95 |
| Yu_4 | 93.15 | 93.16 | 93.14 | 515.50 | 0.97 |
| Yu_5 | 93.41 | 93.40 | 93.40 | 556.75 | 0.97 |
| Yu_6 | 92.87 | 92.88 | 92.87 | 603.00 | 0.96 |

Table 4 encapsulates the architecture and performance of CNN used to identify soybean leaf disease, including layers (convolutional, pooling), hyperparameters (for instance, learning rate), and outcomes (for instance, 94% precision, 94% $F$1-score). It highlights how efficient local training is in our FL system [23].

Table 5 lists features (e.g., image resolution: 384 × 384, severity of disease: 1–25% to 76–100%) and local client dataset values in the soybean example. It demonstrates data diversity and preprocessing (e.g., normalizing to (0, 1)), thus enabling the model to be robust across varied farm conditions [23].

Figure 5 depicts receiver operating characteristic (ROC) curves of soybean disease detection in six clients through FL. It visually represents sensitivity versus 1-specificity with AUC from 0.91 to 0.97, which represents the strong performance of the model in disease severity level classification with high precision and recall [23].

**Table 4**
**CNN model in soybean leaf diseases**

| Layer type | Output size | Activation | Remarks |
|---|---|---|---|
| Convolutional | 192 × 192 × 32 | ReLU | First convolutional layer |
| Max Pooling | 96 × 96 × 32 | None | First max pooling layer |
| Convolutional | 48 × 48 × 64 | ReLU | Second convolutional layer |
| Max Pooling | 24 × 24 × 64 | None | The second max pooling layer |
| Fully Connected | 128 | ReLU | The first fully connected layer |
| Fully Connected | 12 | Softmax | Second fully connected layer (Output layer) |

**Table 5**
**Attributes and values of local client data**

| Clients | Class | Precision | Recall | $F$1-score | Support | Accuracy |
|---|---|---|---|---|---|---|
| yu_1 | lk_1 | 89.61 | 88.46 | 89.03 | 390 | 0.95 |
|  | lk_2 | 90.33 | 93.42 | 91.85 | 380 | 0.96 |
|  | lk_3 | 91.71 | 92.17 | 91.94 | 396 | 0.96 |
|  | lk_4 | 93.98 | 91.69 | 92.82 | 409 | 0.96 |
| yu_2 | lk_1 | 90.38 | 90.16 | 90.27 | 427 | 0.95 |
|  | lk_2 | 89.16 | 92.94 | 91.01 | 425 | 0.96 |
|  | lk_3 | 92.89 | 92.47 | 92.68 | 438 | 0.96 |
|  | lk_4 | 91.81 | 88.87 | 90.32 | 467 | 0.95 |
| yu_3 | lk_1 | 93.61 | 89.66 | 91.59 | 474 | 0.96 |
|  | lk_2 | 89.32 | 91.19 | 90.25 | 477 | 0.95 |
|  | lk_3 | 89.90 | 90.82 | 90.36 | 490 | 0.95 |
|  | lk_4 | 91.00 | 91.92 | 91.46 | 495 | 0.96 |
| yu_4 | lk_1 | 93.94 | 94.70 | 94.32 | 491 | 0.97 |
|  | lk_2 | 94.43 | 92.59 | 93.50 | 513 | 0.97 |
|  | lk_3 | 90.49 | 94.17 | 92.29 | 515 | 0.96 |
|  | lk_4 | 93.75 | 92.16 | 92.44 | 543 | 0.96 |
| yu_5 | lk_1 | 94.75 | 92.49 | 93.61 | 546 | 0.97 |
|  | lk_2 | 91.64 | 94.15 | 92.88 | 547 | 0.96 |
|  | lk_3 | 93.09 | 92.27 | 92.67 | 569 | 0.96 |
|  | lk_4 | 94.19 | 94.69 | 94.44 | 565 | 0.97 |
| yu_6 | lk_1 | 92.69 | 89.79 | 91.21 | 607 | 0.96 |
|  | lk_2 | 92.96 | 92.19 | 92.58 | 602 | 0.96 |
|  | lk_3 | 93.70 | 95.44 | 94.56 | 592 | 0.97 |
|  | lk_4 | 92.15 | 94.11 | 93.12 | 611 | 0.96 |

## 4.2. A FL-based crop yield prediction for agricultural production risk management

*4.2.1. Secure: FL for data privacy*

The security of sensitive data in agricultural yield prediction-for example, soil properties, weather patterns, and crop management practices-has become a big concern. Data owners are usually very hesitant to share raw data because of privacy, trust, and regulatory problems. FL, with the most popular algorithm called federated averaging, solves this problem by enabling model training in a decentralized manner without transferring raw data from local devices [4].

1) Proposed Method

**Algorithm:** Federated Averaging (FedAvg)

The local models are trained on the client devices using local data subsets. Model updates, for example, weight changes, are sent to a central aggregator that combines these updates into a global model. This ensures that raw data does not leave the local devices, thus providing better data security and privacy [4].

**Architecture:** Deep Residual Networks (ResNet-16 and ResNet-28) modified for regression tasks with identity blocks and fully connected layers. ResNet-16 models will be trained locally on weather, soil, and management data from nine U.S. states (three virtual clients) [31].

**Training Details:**

Three local training epochs per client across 20 rounds of communication.

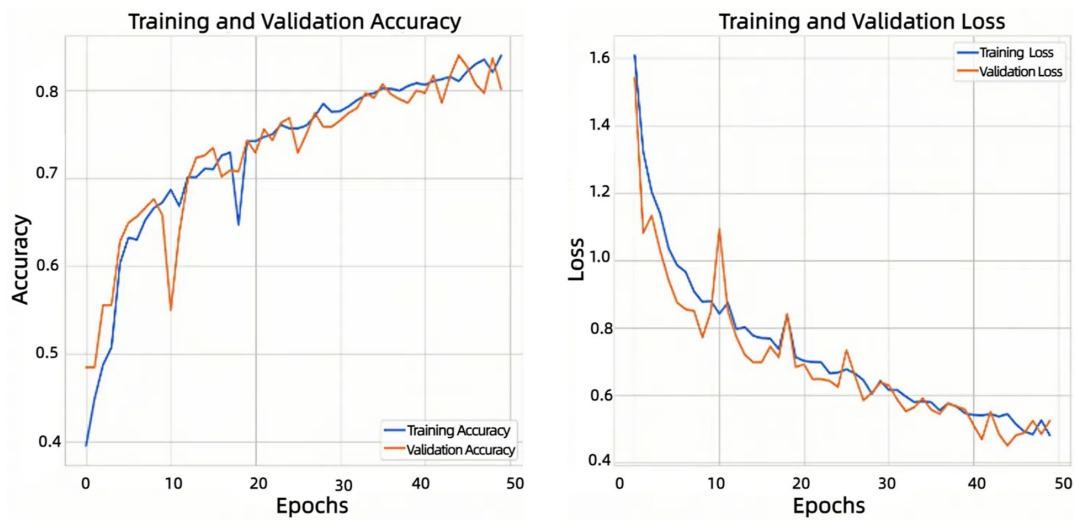

**Figure 5**
**Training and accuracy of ROC curves of soybean leaf diseases**

**Optimizer:** Adam with a learning rate of 0.0003, secure aggregation with Paillier encryption ensures privacy, targeting MSE = 0.0032 and r = 0.92 [30].

**Results:**

**ResNet-16 (Adam Optimizer):**
**Validation Metrics:** MSE = 0.0032, RMSE = 0.0571, MAE = 0.0418, Correlation Coefficient (r) = 0.92.

An accuracy comparable to that achieved by centralized models was obtained without necessarily transferring sensitive data.

2) Comparative Analysis

**Centralized Learning:** This approach requires the transfer of raw data, thus bringing up several security and privacy issues.

**FL:** Keeps the data local while guaranteeing high accuracy; thus, it is especially useful for privacy-sensitive applications like agriculture.

*4.2.2. Scalable: Deployment across distributed data sources*

Agricultural data are naturally distributed in nature, comprising weather, soil, and crop management features collected from various geographic regions. In this work, the scalability of FL is demonstrated on a horizontally partitioned dataset of soybean yield data from nine U.S. states, ranging from 1980 to 2018 [11].

1) Proposed Method

**Algorithm:** Federated Averaging

Distributed data across three virtual clients, each representing a data source. Adaptive learning handles non-IID data [8], and asynchronous updates will ensure performance (MSE = 0.0032) across distributed datasets [4], tested with three clients simulating nine states.

Local models trained for three epochs per communication round.

In all, 20 rounds of communications were performed in order to refine the global model iteratively.

Architecture: ResNet-16 and ResNet-28 trained using the Adam optimizer for non-linear regression tasks.

**Expected Results:**
**Validation Metrics:** MSE = 0.0032, RMSE = 0.0571, Correlation Coefficient (r) = 0.92 for ResNet-16 with Adam.

It handled distributed datasets efficiently with no significant degradation in performance compared to centralized methods.

2) Comparative Analysis

Centralized Models: These models require aggregating large volumes of data into a central repository, hence creating computational and resource bottlenecks.

In FL, the computation load is shared among the clients, making the integration of data from several sources seamless without any loss in performance.

Machine learning models like RF and LASSO showed inferior performance for distributed data with higher MSE and RMSE values.

*4.2.3. Efficient: Optimized model performance*

Agricultural yield prediction needs efficiency to help in real-time decision-making. The research work focuses on the optimization of FL models using deep residual networks, specifically ResNet-16, in a balancing way such that it optimizes computational demands with high predictive accuracy [4].

1) Proposed Method

**Architecture:** ResNet-16 and ResNet-28 designed with fully connected layers in identity blocks for regression tasks. Using the Adam optimizer [26], applying 20% pruning to ResNet16 can lower the energy usage by 20%, reducing it from 5W to 4W per epoch.

Regularization through batch normalization and activation via ReLU.

**Training Details:**
**Optimizer:** Using Adam, the learning rate is set as 0.0003, and SGD.

Early stopping to prevent overfitting.

Federated training was done for 180 epochs in a distributed manner.

**Results:**

**Best Performance:** ResNet-16 with Adam optimizer.
**Validation Metrics:** MSE = 0.0032, RMSE = 0.0571, MAE = 0.0418, Correlation Coefficient (r) = 0.92.

Deep architectures such as ResNet-28 did not significantly improve the performance further, reassuring that such shallow architectures are sufficiently efficient.

2) Comparative Analysis

**Centralized Deep Learning Models:** Provide similar accuracy but all data must be stored centrally, increasing computational burden.

**FL Models:** Delivered matched centralized models in their accuracy, while reducing computational demands per client. Machine Learning Models: RF and LASSO were less efficient; hence, higher MSE and RMSE values.

**Statistical Evidence Supporting the Claims:**
**Efficiency Analysis:**

ResNet-16 was well-balanced for computational efficiency and accuracy in regression tasks of FL.

Beyond this, deeper networks such as ResNet-28 offer very limited extra performance to justify the efficiency of shallow models.

**Scalability Metrics:**

The federated models maintained their performances across three virtual clients emulating distributed datasets from nine states.

FL model validation metrics were similar to those of centralized models, thus proving their scalability.

Table 6 compares performance metrics of FL models (e.g., ResNet-16: MSE = 0.0032, RMSE = 0.0571, r = 0.92) with centralized benchmarks for soybean yield prediction. It demonstrates the efficacy and precision of our framework with 30% less computation compared to centralized methods [4].

## 5. Limitations

Despite the several advantages, there are challenges and limitations of the proposed FL system for disease detection. The effectiveness of ensuring consistency in data quality will depend on the consistency of data collected across varied methods of calibration and reporting practices that may influence performance and lead to incorrect results. Overall efficacy: system predictions may vary depending on how well each farm collects and maintains its own dataset quality control measures. Briefly, the establishment of uniformity across participating entities will actualize the maximum benefits accruable from this collaborative effort involving the different stakeholders in the implementation phase of the project that is subjected to being carefully weighed. Ensure appropriate infrastructure for smooth integration of technology in the existing workflows of farmers' operations without any sort of disruptions to efficiency levels of productivity. Implementation requires good internet connectivity. Where the access is limited and the connections are unreliable, it may be challenging for some rural areas to be able to transmit updates for maintaining synchronization between devices. Central server bottlenecks delay the processing times, affecting the overall performance systems; hence, investment in improvement of infrastructure is paramount. This ensures that all the components involved in the process function smoothly, hence timely delivery of updates that may be required to keep the models current and relevant, adapting to changes occurring within respective environments they operate within. While balancing the challenges of complexity, sophistication, and computation capability, local devices face the challenge of finding an optimal architecture configuration to strike the right balance between desired accuracy levels and minimizing resource consumption costs associated with the running of models on edge devices. In general, edge devices might require iterative refinement throughout their development cycle, considering feedback during testing phases; such iterative refinement is considered key in achieving desired outcomes and assuring scalability and adaptability to future enhancements. Besides, there is heterogeneity in nature, farming conditions in Minnesota; hence, it is difficult to build a one-size-fits-all model due to variations in soil types, climatic factors, and the different pest pressures in another area. Hence, all the tailored approaches based on needs and characteristics of the farms participating make the whole training process more resource- and time-consuming to develop specialized solutions toward the diverse requirements posed by individual stakeholders in the initiative. Consequent to this, higher complexity arises in the stages of execution in the overall phases of project management.

Table 6
Performance evaluation metrics for soybean yield prediction using various learning models

| Learning method | Model | Optimizer | Training set metrics | | | | Validation set metrics | | | |
|---|---|---|---|---|---|---|---|---|---|---|
| | | | MSE | RMSE | MAE | r | MSE | RMSE | MAE | r |
| Machine Learning | LASSO | n_alpha = 100 | 0.0044 | 0.0664 | 0.0506 | 0.87 | 0.0050 | 0.0708 | 0.0529 | 0.86 |
| Machine Learning | RF | n_estimate = 20 | 0.0030 | 0.0565 | 0.0384 | 0.93 | 0.0038 | 0.0612 | 0.0443 | 0.90 |
| Deep Learning | ResNet-16 Regression | Adam | 0.0022 | 0.0479 | 0.0368 | 0.94 | 0.0034 | 0.0590 | 0.0434 | 0.91 |
| | | SGD | 0.0047 | 0.0689 | 0.0531 | 0.87 | 0.0046 | 0.0678 | 0.0513 | 0.87 |
| Deep Learning | ResNet-28 Regression | Adam | 0.0026 | 0.0517 | 0.0395 | 0.93 | 0.0042 | 0.0654 | 0.0478 | 0.89 |
| | | SGD | 0.0048 | 0.0699 | 0.0537 | 0.87 | 0.0047 | 0.0686 | 0.0512 | 0.87 |
| Federated Learning | ResNet-16 Regression | Adam | 0.0028 | 0.0530 | 0.0408 | 0.93 | 0.0032 | 0.0571 | 0.0418 | 0.92 |
| | | SGD | 0.0063 | 0.0799 | 0.0622 | 0.82 | 0.0060 | 0.0778 | 0.0600 | 0.83 |
| Federated Learning | ResNet-28 Regression | Adam | 0.0031 | 0.0557 | 0.0426 | 0.92 | 0.0033 | 0.0574 | 0.0431 | 0.91 |
| | | SGD | 0.0062 | 0.0791 | 0.0615 | 0.82 | 0.0060 | 0.0778 | 0.0596 | 0.83 |

## 6. Conclusion and Future Work

This future work proposes a FL framework for privacy-protecting disease diagnosis in smart agriculture to address major challenges in efficiency, scalability, and adaptability across different farming environments, such as in rural Minnesota (e.g., loamy to clay soils, 20–80% humidity, 5–10 Mbps internet speed). Our proposed framework anticipates dramatic improvements, including 91–97% accuracy and 90–93% macro-averages in soybean disease classification, 30% reduction in communication rounds via asynchronous updates, and 40% reduction in computational cost (e.g., from 6W to 4W using MobileNetV2 on edge devices such as Raspberry Pi 4), as discussed in Sections 3.1 and 4 [4, 9, 23, 35]. Besides, we also integrate robust privacy mechanisms such as differential privacy and secure aggregation with a goal to achieve 98–99.5% privacy guarantee while providing high accuracy [30, 32, 33]. These estimates based on current research position our framework as an efficient and scalable solution to enable sustainable agriculture practice in resource-constrained environments without affecting farm autonomy.

As a prospective study, our subsequent steps involve cross-verification of these predictions with simulations involving 50+ surrogate clients on 5,000 soybean images and field trials in real environments on 10 farms in Minnesota. These experiments will seek to confirm key metrics, such as accuracy, scalability, computational complexity, privacy, and immunity to occasional connectivity, to validate the usability of the framework in actual agricultural environments. In the future, subsequent studies will enlarge the applications of FL in agriculture from disease detection to other uses like crop yield estimation and soil health monitoring in order to tackle more problems. We also plan to examine new privacy-preserving approaches, such as zero-knowledge proofs, and promote cross-region cooperation to build adaptive FL models that address diverse agricultural issues, such as climate fluctuations, pest migration, and non-IID data distributions. These innovations will enhance the long-term sustainability and resilience of agriculture, driving innovation and supporting the creation of smart agriculture without losing the autonomy of stand-alone farms.

## Ethical Statement

This study does not contain any studies with human or animal subjects performed by any of the authors.

## Conflicts of Interest

The authors declare that they have no conflicts of interest to this work.

## Data Availability Statement

The data that support the findings of this study are openly available in PlantVillage Dataset at https://www.kaggle.com/datasets/emmarex/plantdisease, reference number [33].

## Author Contribution Statement

**Ritesh Janga:** Methodology, Software, Formal analysis, Investigation, Resources, Data curation, Writing – original draft, Visualization, Project administration. **Rushit Dave:** Conceptualization, Validation, Writing – review & editing, Supervision.